\documentclass[letterpaper, 10 pt, journal, twoside]{ieeetran}
\IEEEoverridecommandlockouts
\usepackage{dsfont}
\usepackage{cite}
\usepackage{amsmath,amssymb,amsfonts}
\usepackage{algorithmic}
\usepackage{graphicx}
\usepackage{textcomp}
\usepackage{xcolor}
\usepackage{tabularx}
\usepackage{graphicx}
\usepackage{amsmath}
\usepackage{subcaption}
\usepackage{amssymb}
\usepackage{booktabs}
\usepackage{multirow}
\usepackage{xspace}
\usepackage{makecell}
\usepackage{pifont}
\usepackage{tcolorbox}
\usepackage{xcolor,colortbl}         % colors
\usepackage{standalone}
\usepackage{booktabs}
\usepackage{array}
\usepackage[numbers]{natbib}
\usepackage{caption}

\usepackage{threeparttable}
\usepackage[pagebackref,breaklinks,colorlinks]{hyperref}

 \usepackage{textcomp}
\usepackage{pgfplots}
\usetikzlibrary{patterns,positioning,arrows,arrows.meta,calc,shapes,pgfplots.groupplots,fit,backgrounds}
 \usepackage{textcomp}
 
\usepackage[capitalize]{cleveref}
\crefname{section}{Sec.}{Secs.}
\Crefname{section}{Section}{Sections}
\Crefname{table}{Table}{Tables}
\crefname{table}{Tab.}{Tabs.}

\newtcolorbox{findingbox}{
    colback=black!8,        % <-- Set background color (8% black, 92% white)
    colframe=black,         % Frame color
    boxrule=0.5pt,          % Frame thickness
    arc=3mm,                % <-- Set corner radius (e.g., 3mm)
}

\usepackage{amsmath,amsfonts,bm}

\def\1{\bm{1}}

\DeclareMathAlphabet{\mathsfit}{\encodingdefault}{\sfdefault}{m}{sl}
\SetMathAlphabet{\mathsfit}{bold}{\encodingdefault}{\sfdefault}{bx}{n}

\definecolor{gray60}{gray}{0.8}
\newcolumntype{g}{>{\columncolor{gray60}}c}
\def\BibTeX{{\rm B\kern-.05em{\sc i\kern-.025em b}\kern-.08em
    T\kern-.1667em\lower.7ex\hbox{E}\kern-.125emX}}

\begin{document}

\title{
\includegraphics[height=1.3em]{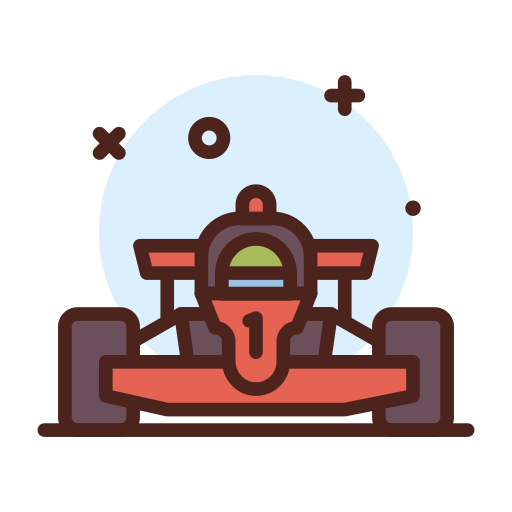} 
PRIX: Learning to Plan from Raw Pixels for End-to-End Autonomous Driving}

\markboth{IEEE Robotics and Automation Letters. Preprint Version. March, 2026}
{Wozniak \MakeLowercase{\textit{et al.}}: PRIX: Learning to Plan from Raw Pixels} 

\author{Maciej Wozniak$^{1}$, Lianhang Liu$^{1,2}$, Yixi Cai$^{1}$, Patric Jensfelt$^{1}$%
\thanks{Manuscript received: Nov 19, 2025; Revised Jan 18, 2026; Accepted March 3, 2026.}%Use only for final RAL version
\thanks{This paper was recommended for publication by Editor A. Bera upon evaluation of the Associate Editor and Reviewers’ comments.
This work was supported by the Wallenberg AI, Autonomous Systems and Software Program (WASP)} %Use only for final RAL version
\thanks{$^{1}$All authors are with the Robotics Perception and Learning Department, KTH Royal Institute of Technology, Stockholm, Sweden. Corresponding Author: Yixi Cai ({\tt\footnotesize yixica@kth.se})}%
\thanks{$^{2} $Second Author is also with SCANIA, Stockholm, Sweden}%
\thanks{Digital Object Identifier (DOI): see top of this page.}
}

\maketitle

%[trim=Left Bottom Right Top, clip]

\maketitle

\begin{abstract}
While end-to-end autonomous driving models show promising results, their practical deployment is often hindered by large model sizes, a reliance on expensive LiDAR sensors and computationally intensive BEV feature representations. This limits their scalability, especially for mass-market vehicles equipped only with cameras. To address these challenges, we propose \textbf{PRIX} (\textbf{P}lan from \textbf{R}aw p\textbf{IX}els). Our novel and efficient end-to-end driving architecture operates using only camera data, without explicit BEV representation and forgoing the need for LiDAR. PRIX leverages a visual feature extractor coupled with a generative planning head to predict safe trajectories from raw pixel inputs directly. A core component of our architecture is the Context-aware Recalibration Transformer (CaRT), a novel module designed to effectively enhance multi-level visual features for more robust planning. PRIX achieves SOTA performance on the NavSim-v2 and nuScenes datasets. On NavSim-v1, it also outperforms the majority of multimodal planners and other camera-only approaches on most metrics. Critically, PRIX is significantly more efficient on NavSim-v1, boasting faster inference speeds and a smaller model size. This combination of performance and efficiency makes it a practical solution for real-world deployment. Our work is open-source and the code will be available upon publication. Check our project website for more  \url{https://maxiuw.github.io/prix}.

% We demonstrate through comprehensive experiments that PRIX achieves state-of-the-art performance on the \textcolor{black}{NavSim-v2 and nuScenes benchmarks, and matches the capabilities of larger, multimodal diffusion planners while being significantly more efficient in terms of inference speed and model size on Navsim-v1}, making it a practical solution for real-world deployment. Our code will be open-source.
%and the code will be at \url{https://maxiuw.github.io/prix}. 
\end{abstract}
\begin{IEEEkeywords}
Deep Learning for Visual Perception, Visual Learning, Motion and Path Planning
\end{IEEEkeywords}

\section{Introduction}

\IEEEPARstart{I}{n} recent years, end-to-end autonomous driving has emerged as a prominent research direction, driven by its ``all-in-one'' training pipeline and goal-oriented output (final trajectory). End-to-end models aim to learn a direct mapping from sensor inputs to the vehicle’s trajectory through large-scale data-driven approaches. Compared with traditional modular pipelines, where perception, prediction, and planning are trained and designed, this paradigm streamlines the overall system and reduces the risk of error propagation between subsystems~\citep{chen2024end}. However, achieving robust and scalable end-to-end solutions in real-world, dynamic environments remains a major challenge.

%Whether using cameras, LiDAR, or both, the computationally intensive process of \textit{feature extraction} remains the primary bottleneck in modern end-to-end architectures.

\textcolor{black}{Whether using cameras, LiDAR, or both, the computational cost of feature extraction significantly constrains the resources available for increasingly complex end-to-end planning architectures.} Current state-of-the-art (SOTA) end-to-end autonomous driving methods~\cite{diffusiondrive,goalflow,li2024hydra,transdiffuser} have focused on fusing multiple sensor modalities, primarily camera and LiDAR, to build a comprehensive environmental representation~\cite{diffusiondrive,goalflow,li2024hydra,transdiffuser,transfuser}. While effective, this reliance on expensive LiDAR sensors and computationally intensive methods limits the scalability of such systems, particularly for mass-market consumer vehicles, which are typically equipped only with cameras, limiting their applicability to vehicles with more expensive sensor suites. Moreover, all these methods depend on BEV (Bird's-Eye View) features, which are computationally expensive, especially for the camera branch, which has to be cast to BEV, e.g., LSS-type models~\cite{lss}. 
On the other hand, many existing camera-only end-to-end approaches suffer from significant practical limitations. Notably, leading camera-only architectures like UniAD and VAD~\cite{uniad,vad} are often oversized, containing over 100 million parameters. This large size makes them computationally expensive, resulting in slower inference speeds and more demanding training requirements. 

\begin{figure}[t!]
    \centering
    \includegraphics[width=.8\linewidth]{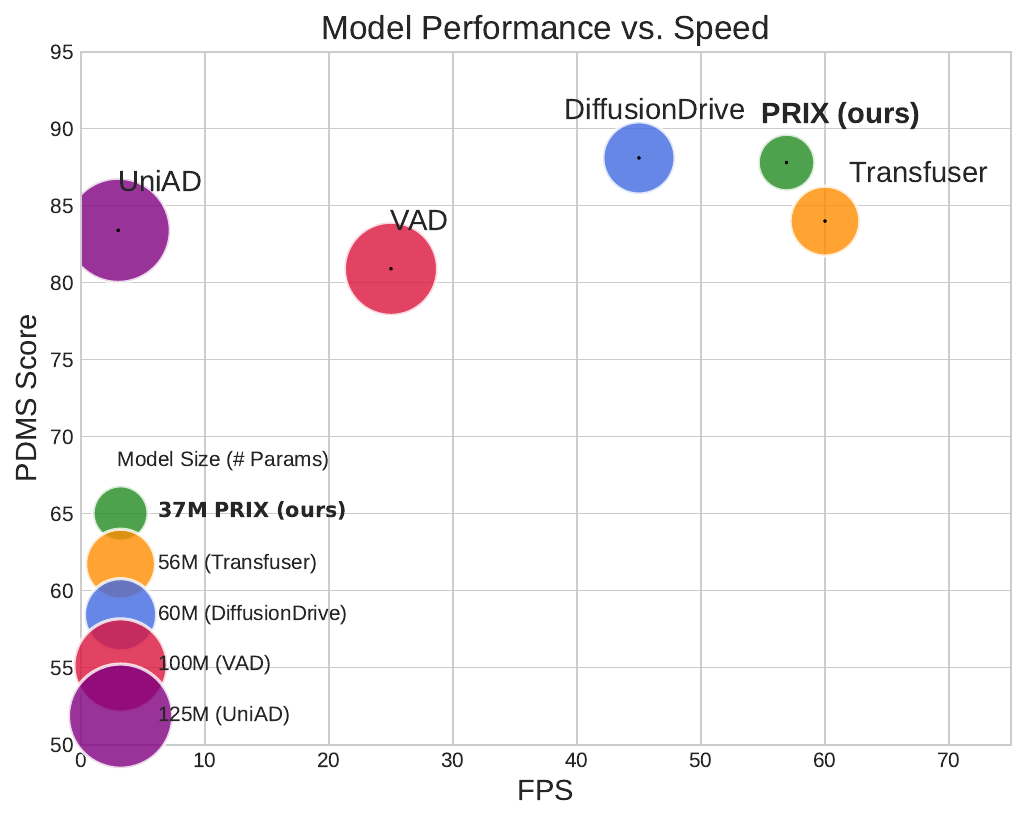}
    \caption{Performance vs. inference speed comparing our camera-only model, \textcolor{green}{\textbf{PRIX}}, to leading methods on the NavSim-v1 benchmark. \textbf{PRIX} outperforms or matches the performance of multimodal methods  SOTA like DiffusionDrive~\cite{diffusiondrive}, while being significantly smaller and faster. Notably, it operates at a highly competitive framerate, falling only 3 FPS behind the fastest model, Transfuser~\cite{transfuser}, while substantially outperforming it in PDMS.}
    \label{fig:firstfig}
\end{figure}

While all components of end-to-end models are integral, we argue that the primary \textit{determinant of system performance} is the visual feature extractor. Its ability to learn task-relevant representation plays the key role in the success of the downstream planning task. However, it is also often the visual feature extractor that is driving the computational cost. 

We posit that it is possible to learn rich visual representations directly from camera inputs for planning without explicitly depending on BEV representation or 3D geometry from LiDAR. Through a detailed analysis of training losses, model design, and experiments with various planning heads, we demonstrate the importance of visual features in end-to-end learning. Our focus on visual camera-only learning is motivated by recent advancements from visual foundation models and world models~\cite{worldnavigation} that have proven that rich, high-fidelity 3D representations of the world can be learned directly from cameras~\cite{hess2025splatad,tonderski2024neurad}. This camera-only paradigm opens the door for powerful, low-cost autonomous systems suitable for a wide range of customer-level vehicles. The autonomous driving domain is particularly well-suited for this approach; vehicles are commonly equipped with 6 to 10 cameras~\cite{caesar2020nuscenes,navsim1,navsim2}, making learning of spatial visual representation feasible.

Inspired by these works, we propose \textbf{P}lan from \textbf{R}aw P\textbf{ix}els (\textbf{PRIX}): a novel end-to-end driving architecture that operates using only camera data and forgoes the need for LiDAR or BEV features. Our method uses a smart visual feature extractor coupled with a generative planning head to directly predict safe trajectories. We demonstrate that our approach successfully predicts future trajectories outperforming other camera-only and most of the multimodal SOTA approaches while being significantly faster and requiring less memory, as shown in \cref{fig:firstfig}. This makes PRIX a practical solution for real-world deployment. Our contributions are as follows:
\begin{itemize}
    \item We introduce \textbf{PRIX}, a novel camera-only, end-to-end planner that is significantly more efficient than multimodal and previous camera-only approaches in terms of inference speed and model size.
    \item We propose \textbf{Context-aware Recalibration Transformer (CaRT)}, a new module designed to effectively enhance multi-level visual features for more robust planning.
    \item We provide a \textbf{comprehensive ablation study} that validates our architectural choices and offers insights into optimizing the trade-off between performance, speed, and model size.
    \item Our method achieves SOTA performance on the NavSim-v2 and nuScenes datasets, as well as most of the NavSim-v1 metrics outperforming the majority of multimodal planners and other camera-only approaches while being much smaller and faster.
\end{itemize}
\section{Related work}

\begin{figure*}[h]
    \centering
    %[trim=Left Bottom Right Top, clip]
    \includegraphics[
    trim=0 235 0 0, % Values are in bp (big points)
    clip,
    width=.75\textwidth]{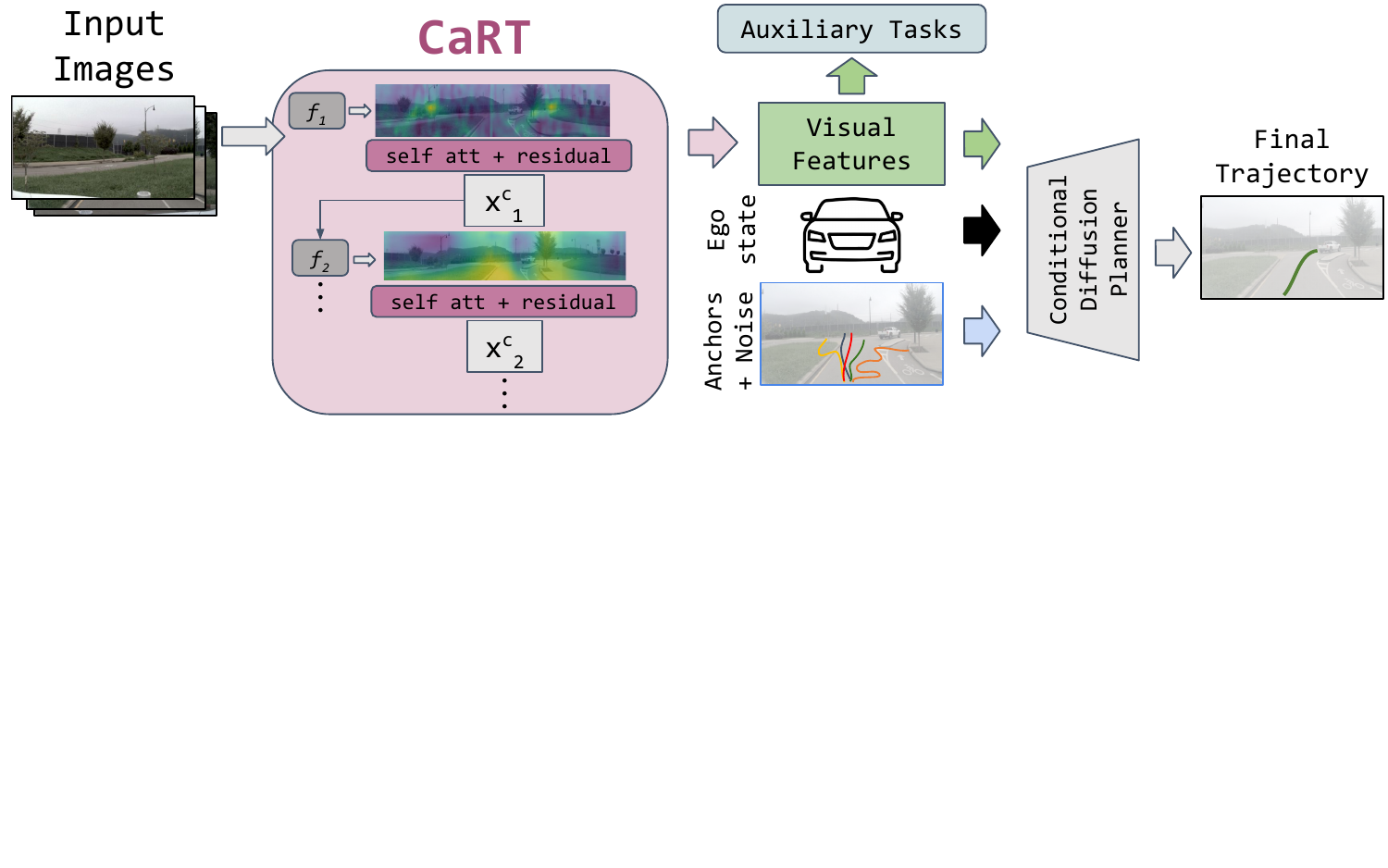}
    \caption{\textbf{PRIX Overview}: Visual features from multi-camera images are extracted by ResNet layers ($f_i$) and together with self-attention and skip connections (\textit{CaRT}, described in \cref{sec:visual}). Next, visual features are used for auxiliary perception tasks (see \cref{sec:loss}) and trajectory planning (see \cref{sec:diff}). 
    }
    \label{fig:main}
    \vspace{-.3cm}
\end{figure*}
% A conditional diffusion planner then uses visual features, along with the current ego state and a set of noisy anchors, to generate the final output trajectory.
\noindent
\textbf{Multimodal End-to-End Driving } To achieve a comprehensive perception of the environment, many recent studies emphasize fusing data from multiple sensors like cameras and LiDAR \cite{wozniak2023toward}. Initial works like Transfuser~\cite{transfuser} used a complex transformer architecture for this fusion. Building this robust world model is the foundational first step; however, the ultimate goal is to translate this perception into safe and effective driving actions. This crucial transition from perception to planning has spurred its own wave of innovation. Early approaches like Hydra-MDP~\cite{li2024hydra} discretized the planning space into sets of trajectories. To overcome the limitations of predefined anchors (pre-set potential trajectories), subsequent research has focused on generating more flexible, continuous paths. This includes diffusion models like TransDiffuser~\cite{transdiffuser} or DiffusionDrive~\cite{diffusiondrive}, which create diverse trajectories without thousands of anchors. To further reduce inference complexity, GoalFlow~\cite{goalflow} employs a flow matching method instead of diffusion, which learns a simpler mapping from noise to the trajectory distribution. 

Architectural innovations have also been key; such as DRAMA~\cite{yuan2024drama} that leverages the Mamba state-space model for computational efficiency or ARTEMIS~\cite{feng2025artemis} uses a Mixture of Experts (MoE) for adaptability in complex scenarios

An alternative paradigm is Reinforcement Learning (RL), where models like RAD~\cite{gao2025rad} are trained via trial and error in photorealistic simulations built with 3D Gaussian Splatting, helping to overcome the causal confusion issues of imitation learning. Despite these advances, a critical perspective from Xu et al.~\cite{xu2024towards} highlights a significant performance gap when models are applied to noisy, real-world sensor data, underscoring the importance of robust intermediate perception. 

While SOTA methods demonstrate powerful capabilities, they are often complex and depend on multimodal sensors. In contrast, our proposed method is designed for simplicity, using only a single modality while achieving better or comparable performance.

\noindent
\textbf{Camera only End-to-End Driving}
End-to-end autonomous driving has evolved from camera-only systems to language-enhanced models. Early camera-only methods like UniAD~\cite{uniad} established unified frameworks for perception, prediction, and planning. To improve efficiency over dense BEV representations, subsequent works introduced more structured alternatives, such as the vectorized scenes in VAD~\citep{vad,chen2024vadv2}, sparse representations in Sparsedrive~\cite{sparsedrive}, 3D semantic Gaussians~\cite{hess2025splatad}, or lightweight polar coordinates. Planning processes were also refined through iterative techniques in models like PPAD~\cite{chen2024ppad}, while others focused on robustness with Gaussian processes (RoCA~\cite{yasarla2025roca}) or precise trajectory selection (DriveSuprim~\cite{yao2025drivesuprim}, GTRS~\cite{li2025generalized}). Efficiency has also been addressed at the input level with novel tokenization strategies~\cite{ivanovic2025efficient}.

More recently, Vision Language Models (VLMs) have been integrated to enhance reasoning. LeGo-Drive~\cite{paul2024lego} uses language for high-level goals, while SOLVE~\cite{chen2025solve} and DiffVLA~\cite{jiang2025diffvla} leverage VLMs for action justification and to guide planning. To manage the high computational cost, methods like DiMA~\cite{hegde2025distilling} distill knowledge from large models into more compact planners. The capabilities of these advanced models are assessed using new evaluation frameworks like LightEMMA~\cite{qiao2025lightemma}.

In contrast to many oversized and slower camera-only methods, PRIX is designed to balance high performance with computational speed, as shown in \cref{fig:firstfig}. As shown in \cref{sec:experiments}, our model outperforms other camera-only models on available benchmarks while being much more efficient.

% \noindent
% \textbf{Generative Planning}
% Early end-to-end methods often regressed a single trajectory, which can fail in complex scenarios with multiple valid driving decisions. To address this, recent work has shifted towards generating multiple possible trajectories to account for environmental uncertainty.

% More recently, generative models have become a pivotal tool. DiffusionDrive~\cite{diffusiondrive} applies diffusion models to trajectory generation, introducing a truncated diffusion process to make inference feasible in real-time. In parallel, DiffusionPlanner~\cite{diffusionplanner} leverages classifier guidance to inject cost functions or safety constraints into the diffusion process, allowing the generated trajectories to be flexibly steered. To further reduce inference complexity, GoalFlow~\cite{goalflow} employs a flow matching method, which learns a simpler mapping from noise to the trajectory distribution. Lately, TransDiffuser~\cite{transdiffuser} proposed to combine both anchors and end-points. Inspired by the speed and performance of these methods, generative trajectory heads seems to be a \textit{go-to} approach yielding the best results~\cite{lihydra}
% While generative methods have significantly advanced the field, they are often designed to operate on multi-sensor features. Our work builds upon the insights of generative planning but adapts them to a more efficient, camera-only architecture.

\section{Method}

The goal of our end-to-end autonomous driving model, shown in \cref{fig:main}, is to generate the best future trajectory of the ego-vehicle from raw camera data. We use only the current time-step camera and the ego vehicle’s status (velocity, acceleration, and navigation commands) as an input. Our model outputs: 8-waypoint trajectory over 4 sec at 2 Hz, with each waypoint defined by x, y, and heading. Camera only feature extraction, detailed in \cref{sec:visual}, is a base for the conditional denoising diffusion planner, described in \cref{sec:diff}. We detail and justify our design choices in \cref{sec:design} and the main objective and auxiliary tasks are discussed in \cref{sec:loss}.

\subsection{Visual Feature Extraction}
\label{sec:visual}
\begin{figure}[t]
    \centering
    %[trim=Left Bottom Right Top, clip]
    \includegraphics[
    trim=0 105 85 0, % Values are in bp (big points)
    clip,
    width=.95\linewidth]{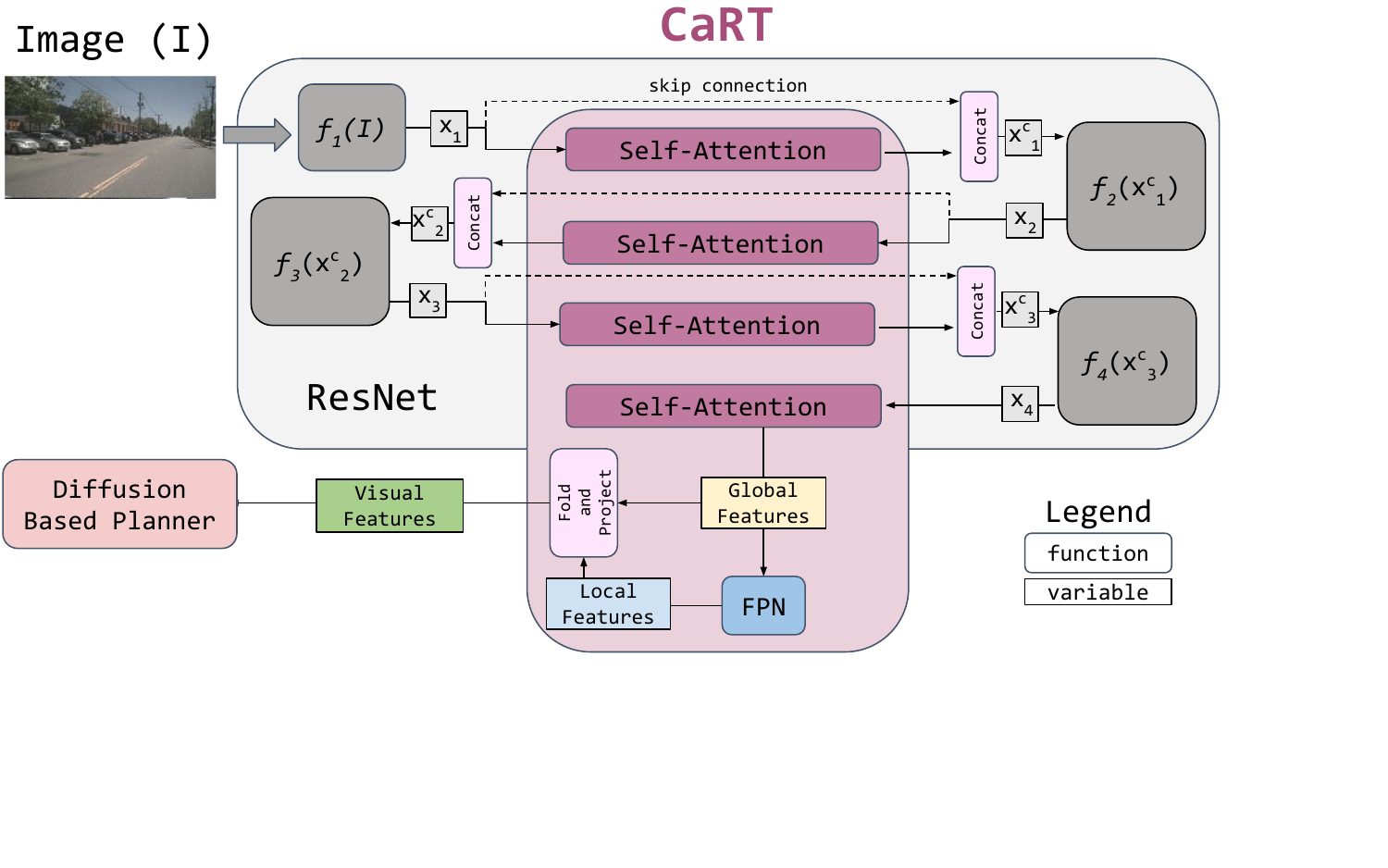}
    \caption{Architecture of our visual feature extractor with  \textbf{Context-aware Recalibration Transformer (CaRT)} module.}
    \label{fig:cart_flow}
\end{figure}

The foundation of our proposed method is a lightweight, camera-only, visual feature extractor designed to derive a rich, \textit{multi-scale representation} of the driving scene, as shown in \cref{fig:cart_flow}. This hierarchical approach is critical for autonomous driving, a task that demands both high-level semantic understanding (e.g., an upcoming intersection) and precise low-level spatial detail (e.g., tracking the exact lane curvature).

To generate and refine these multi-scale features, we employ a ResNet as the hierarchical backbone, which naturally extracts feature maps ($x_i$) at distinct resolutions. However, with raw ResNet features, we face a classic dilemma: early layers capture fine spatial details but lack scene-level understanding, while deeper layers possess rich semantic context but are spatially coarse. \textcolor{black}{To address this, we introduce our novel \textbf{Context-aware Recalibration Transformer (CaRT)} module inspired by previous works as SE \cite{hu2018squeeze} and CBAM~\cite{woo2018cbam}.}

The feature map $x_i$, where $i \in \{1, 2, 3, 4\}$, is first spatially standardized via adaptive average pooling to a fixed size (512 feature size in our implementation, see \cref{sec:design} for ablation studies). Next, features are processed by a self-attention (SA) part of a CaRT module to model long-range dependencies across the spatial domain (see \cref{fig:cart_flow}). A single, weight-shared multi-head self-attention block is applied to each sequence of tokens (explained in \cref{sec:design}). For each feature level $i$, we compute the Query ($Q_i$), Key ($K_i$), and Value ($V_i$) matrices using shared linear projection matrices $W_Q$, $W_K$, and $W_V$: $Q_i = x_iW_Q, K_i = x_iW_K,  V_i = x_iW_V$. 
    % \quad \text{for } i \in \{2, 3, 4, 5\}$

The output of the CaRT module is the attention $A_i$  computed using the scaled dot-product attention $\text{A}(Q_i, K_i, V_i) = \text{softmax}\left(\frac{Q_iK_i^T}{\sqrt{d_k}}\right)V_i$. $A_i$, which is our recalibrated feature map, is then upsampled to the original dimensions of $x_i$, concatenated with the original $x_i$ feature map (extracted from ResNet) via skip connection, creating $x^c_i$, and fed to the next ResNet layer $f_{i+1}$ as shown in \cref{fig:cart_flow}. 

% \textcolor{black}{explain recalibration process what we understand by it}

% The iterative \textit{recalibration} process is actively refining the initial feature maps from the ResNet backbone by infusing them with global semantic context learned via SA as an act of adjusting the value and significance of the initial local features based on this newly understood global context. 
The iterative \textit{recalibration} process actively refines the initial feature maps from the ResNet backbone by infusing them with global semantic context learned via SA. This effectively adjusts the value and significance of the initial local features based on the newly understood global context. It is not just adding new information; it is fundamentally changing the interpretation of the existing features by infusing them with the global context of the entire scene generated by the CaRT self-attention layers. The final feature map is \textit{Global Features}, $x^C_5\in\mathds{R}^{B\times C_c\times H_c\times W_c}$, which encapsulates information from all levels.  
To synthesize the final multi-scale representation, the architecture ends in a top-down pathway, analogous to a Feature Pyramid Network (FPN). The Semantic Features are passed through a series of upsampling and 3x3 convolutional layers to restore a higher-resolution feature map, ensuring it benefits from semantic context while retaining precise spatial understanding, we refer to it as \textit{Local Features}, $x^L_5\in\mathds{R}^{B\times C_L\times H_L\times W_L}$.

\subsection{From Camera Features to Trajectories Without Geometric BEV}
\label{sec:view_transform}

\textcolor{black}{Planning operates on two learned representations: a \textit{Token Memory} (derived from global features $x^C_5$ plus an ego-status token) and a \textit{Planner Grid} (obtained by mixing this memory with local features $x^L_5$). The planner grid is \emph{not} a geometric BEV; it is a learned canonical grid aligned to the ego frame through supervision. We rely on a fixed camera rig with constant view ordering and resolution.}

\textcolor{black}{Under this setup, camera geometry is constant and absorbed by the network parameters. Consequently, the planner grid is anchored to the ego frame solely through semantic and trajectory losses; no camera intrinsics or extrinsics are used.}

\textcolor{black}{The \textit{Token Memory} is built by flattening $x^C_5$ into $(H_cW_c)$ visual tokens and appending a status token. A learned index embedding is added to encode the view identity and the spatial slot. The resulting memory serves as Keys and Values for the transformer decoding stage, where a learned query attends to the token memory before splitting into trajectory and agent queries. To provide spatial structure for the grid, this same memory is ``folded'' back into 2D. The status token is discarded, and the remaining visual tokens are reshaped into an $H_c \times W_c$ map, upsampled to $(H_L, W_L)$, and concatenated with the local feature $x^L_5$. A point-wise projection $\Phi$ then produces the \textit{planner grid} $\mathbf{G}$. The entire fold-and-project pipeline is learned, allowing supervision to stabilize $\mathbf{G}$ as a canonical planning representation.}

%An input feature map $f_i$ is processed in parallel through a skip connection and a recalibration path. The recalibration path uses adaptive pooling and self-attention block to capture global context. The resulting features are upsampled and added back to the original map via a residual connection, producing a refined output that is enhanced with contextual information.
\begin{figure}[t]
    \centering
        %[trim=Left Bottom Right Top, clip]

    \includegraphics[
    trim=0 265 0 0, % Values are in bp (big points)
    clip,
    width=.95\linewidth]{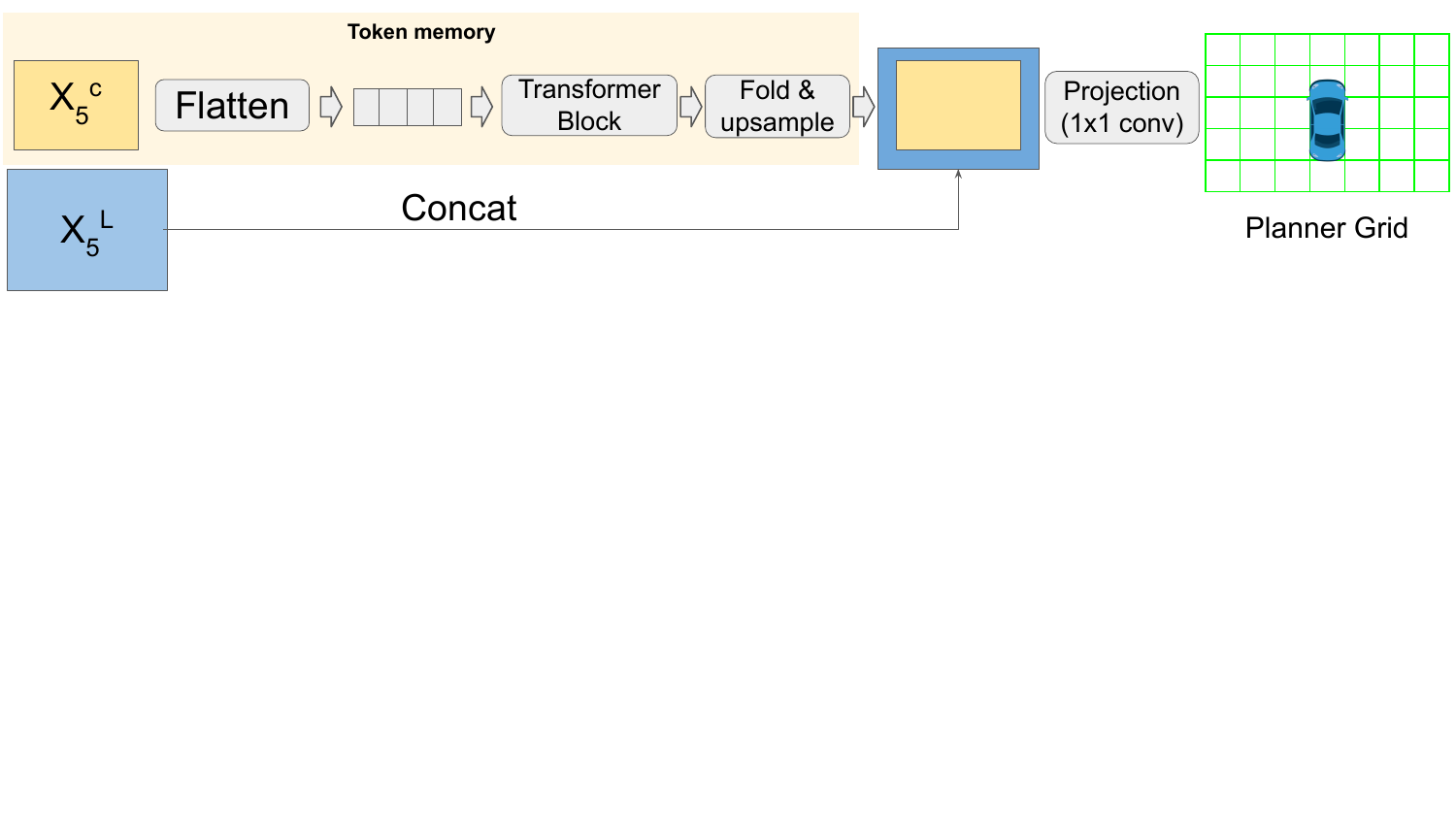}
    \vspace{-.1cm}
    \caption{\textcolor{black}{Fold and Project flowchart.}}
    \label{fig:placeholder}
    \vspace{-.3cm}
\end{figure}

\subsection{Diffusion-Based Trajectory Planner}
\label{sec:diff}
% \textcolor{black}{reduce since not our contrib}
\noindent

For planning, we adopt a conditional denoising diffusion head from DiffusionDrive \cite{diffusiondrive} that generates trajectories via iterative refinement (we also test different planners in \cref{sec:ablations}, showing our method performs well with any planner). Unlike regression-based planners, this approach treats trajectory prediction as a denoising process: given noisy trajectory proposals (anchors), ego state, and visual features, the model gradually refines them into feasible plans.

The trajectory is represented as a sequence of waypoints, $\tau = {(x_t, y_t)}_{t=1}^{T_f}$, where $T_f$ is the planning horizon and $(x_t, y_t)$ is the waypoint location at future time $t$ in the ego-vehicle's coordinate system.

The \textit{forward process}, $q$, progressively adds Gaussian noise to a clean trajectory $\tau^0$ over $n$ timesteps. In one step: $q(\tau^i | \tau^0) = \mathcal{N}(\tau^i; \sqrt{\bar{\alpha}^i}\tau^0, (1-\bar{\alpha}^i)\mathbf{I})$, where $i$ is the diffusion timestep and $\bar{\alpha}^i = \prod_{s=1}^{i} (1-\beta^s)$. As $i \to n$, $\tau^i$ converges to an isotropic Gaussian. The \textit{reverse process} learns to remove noise and recover the trajectory. We train a model, $\epsilon_\theta$, to predict the noise component $\epsilon$ at timestep $i$.

\textcolor{black}{This process is conditioned on visual features $c_{\text{visual}}$, ego state $c_{\text{ego}}$ and a set of noisy anchors. Following DiffusionDrive~\cite{diffusiondrive}, we generate a vocabulary of trajectory anchors by performing K-Means clustering on the ground-truth trajectories of the training dataset. Each cluster centroid represents a distinct driving intention (e.g., turn left, lane keep). During training, we select the anchor closest to the ground truth and corrupt it with Gaussian noise to create $c_{\text{anch}}$. This ``anchored'' initialization provides a strong prior, allowing the diffusion model to refine the trajectory in very few steps (e.g., $n=2$) compared to generating from pure noise, significantly reducing inference latency.}

% This process is conditioned on a context vector, \textit{c}, combining visual features $c_{\text{visual}}$, ego state $c_{\text{ego}}$, and noisy anchors $c_{\text{anch}}$: $c = \text{combine}(c_{\text{visual}}, c_{\text{ego}},c_{\text{anch}})$. Starting from noisy anchors $\tau^n$, we iteratively apply $\epsilon_\theta$ to denoise guided by $c$, yielding context-appropriate trajectories $\tau^0$, from which we select the highest-confidence one. Note that while large $n$ is common in generative models, it increases latency and, as shown in \cref{sec:design}, often drives the model toward simpler, suboptimal solutions.

% \textcolor{black}{rev comment: R1 - Clarify no

\subsection{Design choices and findings}
\label{sec:design}
\noindent
Our initial design consisted of a visual feature extractor with separate SA modules in CaRT corresponding to each feature level of ResNet backbone and two-step diffusion planner. Throughout this section, we analyze our design in detailed ablation studies on Navsim-v1 and evaluate using PDMS (explained in \cref{sec:experiments}) to arrive at the final configuration of our model. 

\noindent
\textit{a) Module Integration Strategy}
Our experiments show that using a CaRT module where the self-attention layers share weights across all feature scales of the backbone outperforms using separate, specialized SA for each $x_i$. As detailed in \cref{tab:ablation_sharing}, this shared-weight design not only achieves a higher score but also reduces the parameter count and increases inference speed. This indicates that the core logic of using global context to recalibrate local features is a universal principle. Forcing a single set of self-attention weights to learn this logic across different levels of feature abstraction results in a more robust and generalized representation.

\begin{table}[h!]
\centering
\caption{\small Ablation on sharing weights in SA layers in CaRT module across different scales.}
\label{tab:ablation_sharing}
\vspace{-.2cm}
\resizebox{.85\columnwidth}{!}{
\begin{tabular}{lccc}
\toprule
\textbf{Configuration} & \textbf{Params} $\downarrow$& \textbf{PDMS $\uparrow$} & \textbf{FPS} $\uparrow$ \\
\midrule
Separate SA & 39M & 87.3& 54.4\\
Shared SA 256 & \textbf{33M} & 87.0& \textbf{57.9}\\
\rowcolor{gray!20} Shared SA 512 & 37M & \textbf{87.8}& 57.0\\
Shared SA 768 & 39M & 87.7 & 56.0  \\
\bottomrule
\end{tabular}
}
% \vspace{-.5cm}
\end{table}

\noindent
\textit{b) Anchors with end points}
Inspired by the concept of GoalFlow~\cite{goalflow}, in \cref{tab:anchors_endpoints} we experimented with using the final end point as an additional conditioning signal for our diffusion head planner, aiming to help the final trajectory objective. We hypothesized that this would complement the guidance from the anchors. However, our findings indicate that the combination of anchors and end points is counterproductive and appears to confuse the planner, creating a conflict between the local, step-by-step guidance from anchors and the global pull of the final destination. As a result, this combination led to a slight degradation in performance, suggesting that anchors alone are a better approach, which we used in our model.

\begin{table}[h!] \centering \caption{\small Ablation on anchors plus end points} \label{tab:anchors_endpoints} 
\renewcommand{\arraystretch}{0.8} % Default value: 1
\setlength{\tabcolsep}{5pt}
\vspace{-.2cm}
\begin{tabular}{c | cc | c} 
\textbf{Model} & Anchors & End-Points & \textbf{PDMS $\uparrow$} 
\\ \midrule 
\rowcolor{gray!20} \textcolor{black}{anchors-only} & \checkmark & & \textbf{87.8 }\\
 \textcolor{black}{end-points only} & & \checkmark  & 83.5\\
 \textcolor{black}{achnors+end points} & \checkmark & \checkmark  & 85.9\\
\end{tabular} 
% \vspace{-.5cm}
\end{table}

\noindent
\textit{c) Overall Impact of CaRT}
To quantify the contribution of the CaRT module and justify its computational cost, we created a baseline version of PRIX without it. The residual connection still exists but processes features that are only downsampled and upsampled, without any transformer-based processing. In \cref{tab:ablation_existence} we show that removing the module reduces parameters and increases speed but model performance drastically drops. Therefore, we included the CaRT module in our final model, as it provides a significant performance boost while remaining highly efficient.

\begin{table}[h!]
\centering
\caption{\small Ablation on the existence of the CaRT module.}
\label{tab:ablation_existence}
\vspace{-.2cm}
\resizebox{.85\columnwidth}{!}{
\begin{tabular}{lccc}
\toprule
\textbf{Configuration} & \textbf{Parameters}$\downarrow$ & \textbf{PDMS $\uparrow$} & \textbf{FPS}$\uparrow$\\
\midrule
\rowcolor{gray!20} PRIX (\textbf{with} CaRT) & 37M & \textbf{87.8} & 57.0\\
PRIX (\textbf{no} CaRT) & \textbf{20M} & 76.4 & \textbf{70.9}\\
\bottomrule
\end{tabular}
}
% \vspace{-.5cm}
\end{table}

\begin{figure}[h]
    \centering
    \includegraphics[width=.8\linewidth]{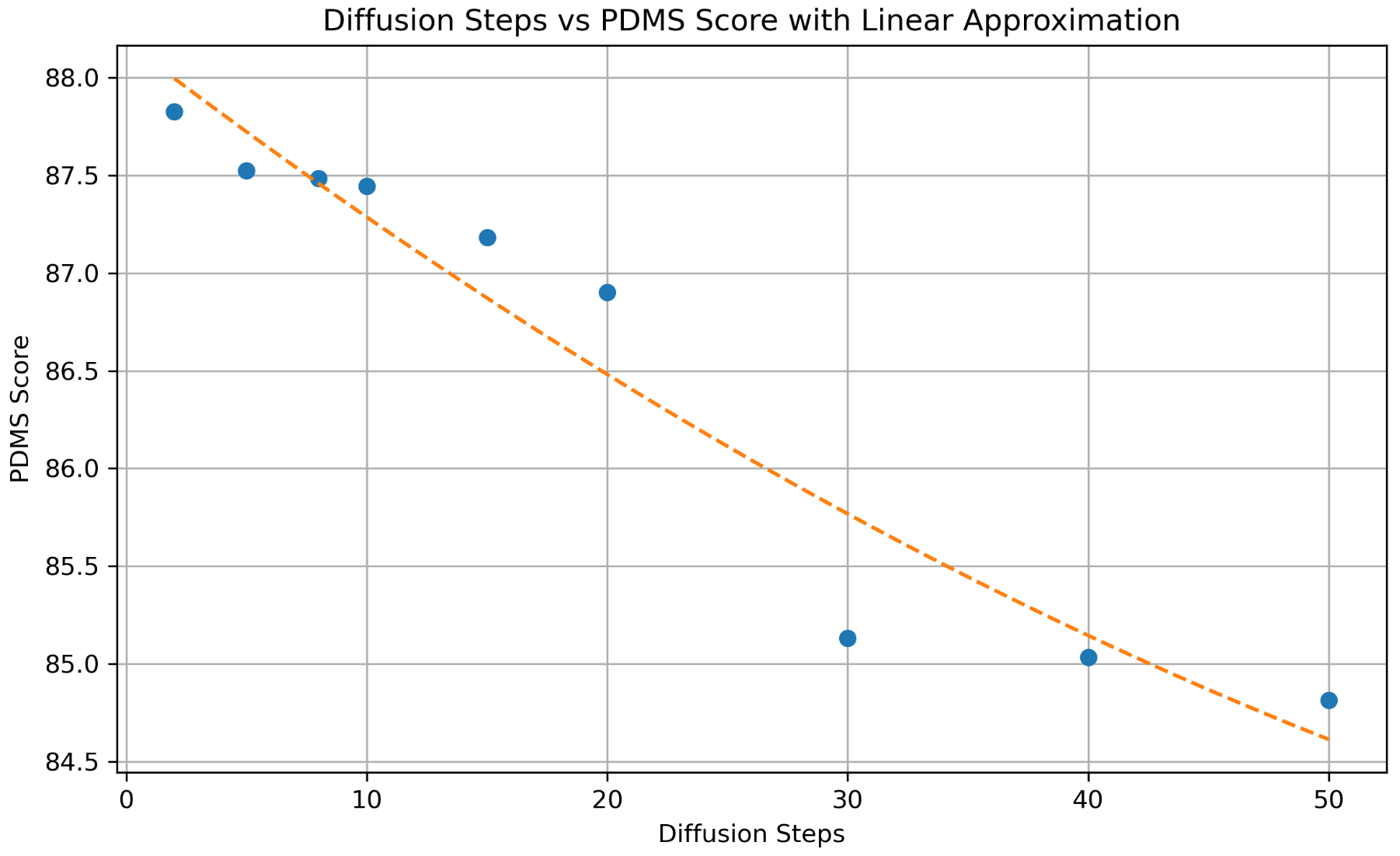}
    \vspace{-.2cm}
    \caption{Diffusion steps vs performance on Navsim-v1.}
    \label{fig:diffusionsteps}
    \vspace{-.3cm}
\end{figure}

\noindent
\textit{d) Diffusion steps} We experimented with various truncated diffusion time steps, specifically 2-50 and evaluated performance using the PDMS shown in \cref{fig:diffusionsteps}. The results showed that performance degrades when the number of diffusion steps increases. Such over-smoothing diminishes the quality of the final predictions, reflected in the notable drop in PDMS at higher step counts; thus, we opt for 2 steps.

\subsection{Training Objectives and Auxiliary Tasks}
\label{sec:loss}

Relying solely on a trajectory imitation loss, as shown in~\cref{tab:loss_ablation} and other works~\cite{transfuser,vad,diffusiondrive}, is \textit{insufficient} for an end-to-end model to learn the rich representations needed for robust autonomous driving. To address this, we employ a multi-task learning paradigm. By adding auxiliary tasks, we introduce a powerful inductive bias that compels our camera-only feature extractor to learn a more structured, semantically meaningful world representation specialized for autonomous driving, ultimately leading to better planning. Our total loss is a weighted sum of the primary planning task and auxiliary objectives:

\begin{equation}
    \mathcal{L} = \lambda_{\text{plan}}\mathcal{L}_{\text{plan}} + \lambda_{\text{det}}\mathcal{L}_{\text{det}} + \lambda_{\text{sem}}\mathcal{L}_{\text{sem}},
    \label{eq:total_loss}
\end{equation}
where $\lambda$ terms are the corresponding loss weights. 

% Detailed architecture of the segmentation and detection heads can be found in the supplementary.

\noindent
\textbf{Primary Planning Loss ($\mathcal{L}_{\text{plan}}$)} Our model learns the ego-vehicle's future path by minimizing the L1 distance between the predicted waypoints $\hat{\mathbf{p}}_{1:T}$ and the ground-truth trajectory $\mathbf{p}_{1:T}$. This loss, defined as $\mathcal{L}_{\text{plan}} = \frac{1}{T}\sum_{t=1}^{T} \left\lVert \hat{\mathbf{p}}_{t} - \mathbf{p}_{t} \right\rVert_{1}$, optimizes the final trajectory.
\noindent
\textbf{Auxiliary Task: Object Detection ($\mathcal{L}_{\text{det}}$)} Safe navigation requires awareness of other road users. We add an auxiliary objective to localize traffic participants like vehicles and pedestrians. This ensures the model's internal representations are sensitive to dynamic agents that influence planning. The detection loss, $\mathcal{L}_{\text{det}} = \lambda_{\text{cls}}\mathcal{L}_{\text{cls}} + \lambda_{\text{reg}}\mathcal{L}_{\text{reg}}$, combines a focal loss for classification and an L1 loss for 3D bounding box regression.
\noindent
\textbf{Auxiliary Task: Semantic Consistency ($\mathcal{L}_{\text{sem}}$)} To ensure the model understands the static driving environment, we introduce a semantic consistency loss. This provides dense, pixel-level supervision, compelling the feature extractor to learn the scene's structure, such as drivable areas and lane boundaries. We apply a pixel-wise cross-entropy (CE) loss, $ \mathcal{L}_{\text{sem}} = \text{CE}(\hat{\mathbf{S}}, \mathbf{S})$, between the predicted $\hat{\mathbf{S}}$ and ground-truth $\mathbf{S}$ semantic maps. This contextual understanding enables more feasible and appropriate trajectories.

\begin{table*}[ht!]
\caption{\small Performance of different models on \textbf{Navsim-v1}. The up arrow \textbf{($\uparrow$)} indicates that \textbf{higher values are better}. Best results are in \textbf{bold}, and second best are \underline{underlined}. C\&L refers to Camera and LiDAR. $\dagger$Default GoalFlow uses V2-99, but they also reported Resnet34. We were only able to experiment with the inference speed of the models that released their code.}
\centering
\label{tab:navsim1_results}
% tighter layout to avoid stretching while fitting width
\setlength{\tabcolsep}{4pt}
\renewcommand{\arraystretch}{1}
\small
\begin{tabularx}{\textwidth}{c c | l c c| c c c c c | c c}
\textbf{Method}  & \textbf{Conference} & \textbf{Input} & \textbf{\# frames} & \textbf{Backbone} & \textbf{NC $\uparrow$} & \textbf{DAC $\uparrow$} & \textbf{TTC $\uparrow$} & \textbf{Comf. $\uparrow$} & \textbf{EP $\uparrow$} & \textbf{PDMS $\uparrow$} & \textbf{FPS $\uparrow$} \\
\hline
% -------- 4 frames (PDMS asc) --------
VADv2~\cite{chen2024vadv2} & \textcolor{black}{arXiv24} & Camera & 4 & Resnet34 & 97.2 & 89.1 & 91.6 & 100 & 76.0 & 80.9 & 25 \\
Hydra-MDP~\cite{li2024hydra} & CVPRW24 & C \& L & 4 & Resnet34 & 97.9 & 91.7 & 92.9 & 100 & 77.6 & 83.0 & — \\
% -------- >=2 / 2 frames (PDMS asc) --------
UniAD~\cite{uniad} & CVPR23 & Camera & $\ge2$ & Resnet34 & 97.8 & 91.9 & 92.9 & 100 & 78.8 & 83.4 & \textcolor{black}{3} \\
PARA-Drive~\cite{paradrive} & CVPR24 & Camera & 2 & Resnet34 & 97.9 & 92.4 & 93.0 & 99.8 & 79.3 & 84.0 & — \\
Hydra-MDP++\cite{lihydra} & CVPRW25 & Camera & 2 & Resnet34 & 97.6 & 96.0 & 93.1 & 100 & \underline{80.4} & 86.6 & — \\
% -------- 1 frame (PDMS asc) --------
% LTF~\cite{transfuser} & TPAMI22 & Camera & 1 & Resnet34 & 97.4 & 92.8 & 92.4 & 100 & 79.0 & 83.8 & — \\
Transfuser~\cite{transfuser} & TPAMI22 & C \& L & 1 & Resnet34 & 97.7 & 92.8 & 92.8 & 100 & 79.2 & 84.0 & \textbf{60} \\
GoalFlow$^{\dagger}$~\cite{goalflow} & CVPR25 & C \& L & 1 & Resnet34 & \textbf{98.3} & 93.8 & \underline{94.3} & 100 & 79.8 & 85.7 & — \\
 DiffusionDrive\cite{diffusiondrive} & CVPR25 & C \& L & 1 & Resnet34 & \underline{98.2} & \underline{96.2} & \textbf{94.7} & 100 & 82.2 & \textbf{88.1} & 45 \\
\hline
\rowcolor{gray!20} \textbf{PRIX (ours)} & RAL26 & Camera & \textbf{1} & Resnet34 & 98.1 & \textbf{96.3} & 94.1 & \textbf{100} & \textbf{82.3} & \underline{87.8} & \underline{57} \\
\end{tabularx}
\end{table*}

\begin{figure*}[ht!]
    \centering
    \begin{subfigure}[t]{0.4\linewidth}
        \centering
        \includegraphics[width=\linewidth]{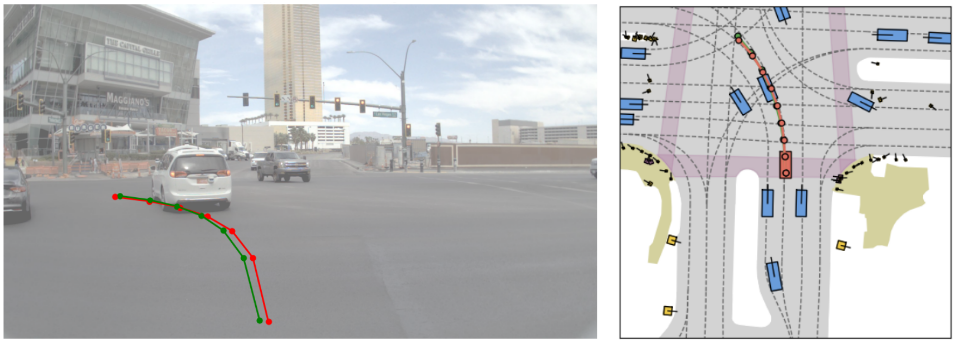}
        \caption{\textcolor{black}{Our model can perform a safe left turn at a busy intersection.}}
        \label{fig:qual5_main}
    \end{subfigure}
    \hfill
    \begin{subfigure}[t]{0.4\linewidth}
        \centering
        \includegraphics[width=\linewidth]{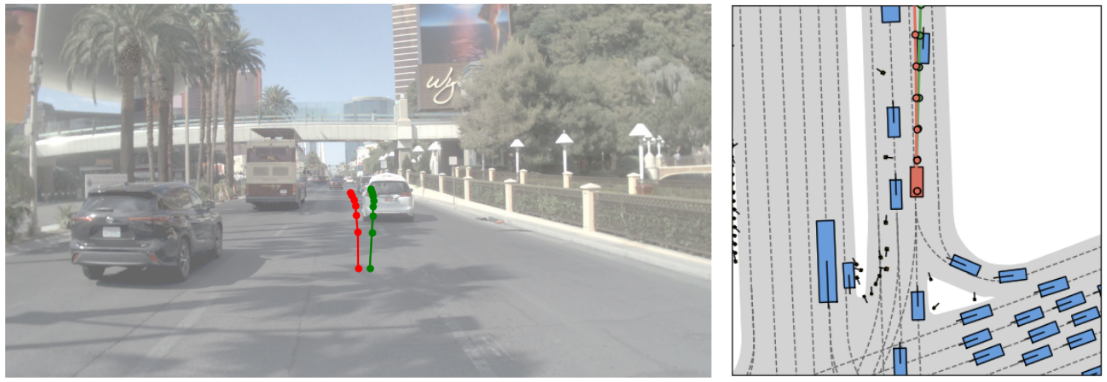}
        \caption{\textcolor{black}{PRIX is safer than GT; it maintains a larger distance from the other car.}}
        \label{fig:qual9_main}
    \end{subfigure}
    \begin{subfigure}[t]{0.3\linewidth}
        \centering
        \includegraphics[width=\linewidth]{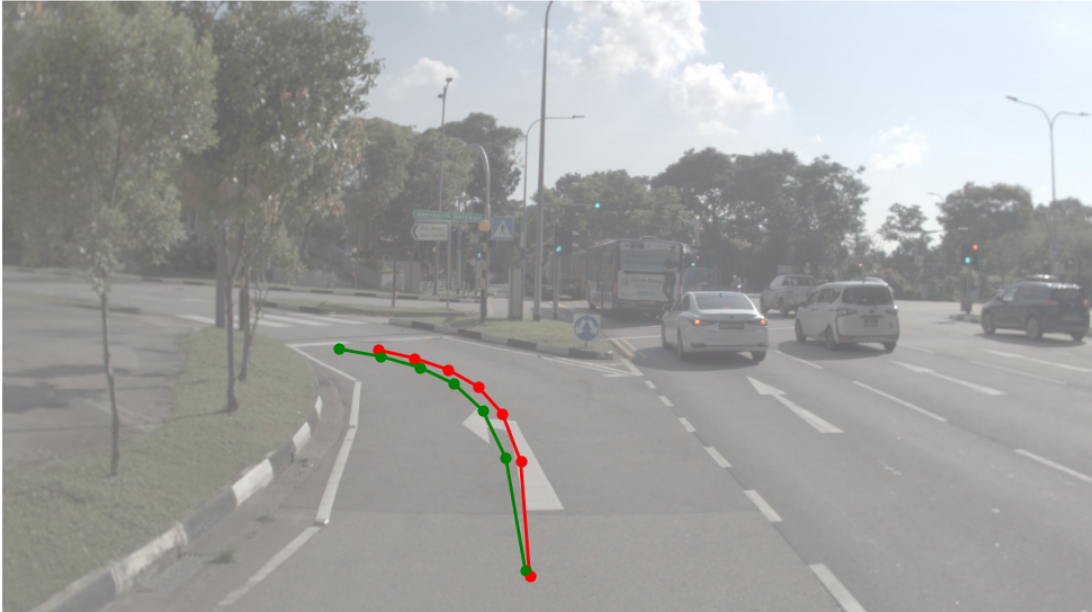}
        \caption{\textcolor{black}{PRIX can navigate well in the good weather,...}}   
        \label{fig:qual5_org}
    \end{subfigure}
    \hfill
    \begin{subfigure}[t]{0.3\linewidth}
        \centering
        \includegraphics[width=\linewidth]{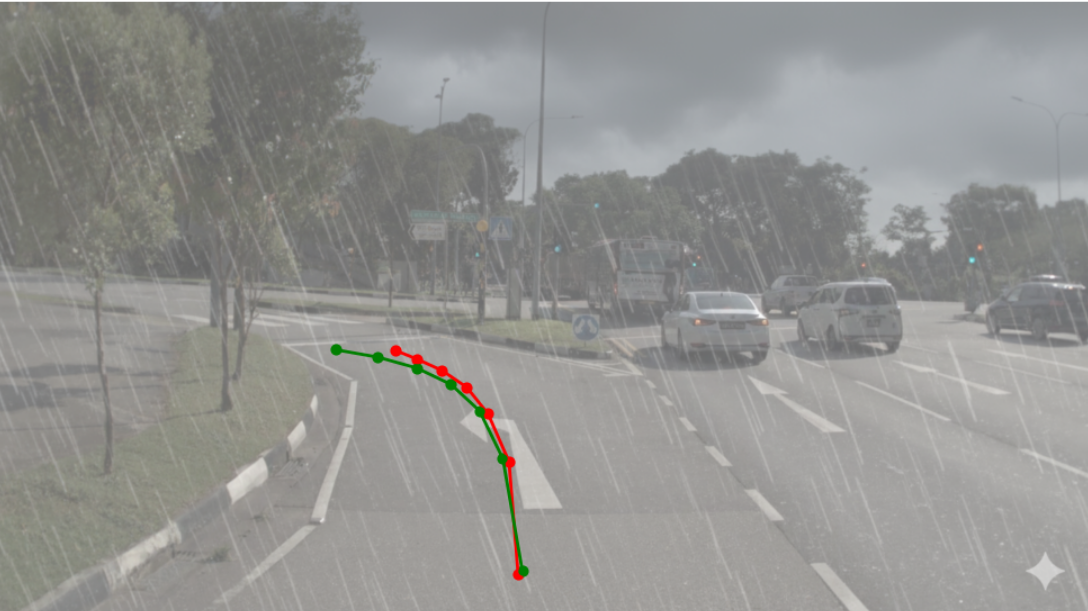}
        \caption{and adverse conditions like rain...}
        \label{fig:qual5_rain}
    \end{subfigure}
    \hfill
    \begin{subfigure}[t]{0.3\linewidth}
        \centering
        \includegraphics[width=\linewidth]{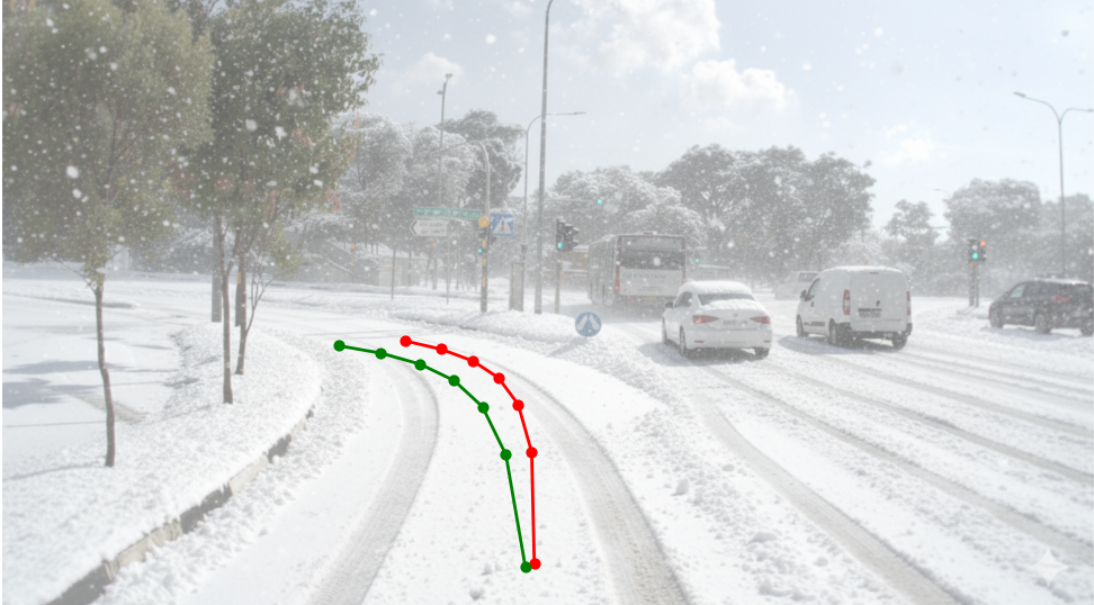}
        \caption{...or snow.}
        \label{fig:qual9_snow}
    \end{subfigure}
    % \vspace{-.3cm}
    \caption{Qualitative trajectory predictions from PRIX. In some cases, like \ref{fig:qual9_main}, our predictions are safer than the GT.}
    \label{fig:qual_combined}
\end{figure*}

\begin{table*}[h!]
\centering
\caption{\small Performance comparison of different driving models for \textbf{Navsim-v2}. The up arrow \textbf{($\uparrow$)} indicates that \textbf{higher values are better}. Best results are in \textbf{bold}, and second best are \underline{underlined}. All the methods are camera-only.}
\label{tab:navsim2_evaluation}
\definecolor{lightgreen}{HTML}{E7F5E8} % Define a color similar to the image
\resizebox{\textwidth}{!}{
\begin{tabular}{lc|cccccccccc}
% \toprule
\textbf{Method} & \textbf{Backbone} & \textbf{NC} $\uparrow$ & \textbf{DAC} $\uparrow$ & \textbf{DDC} $\uparrow$ & \textbf{TL} $\uparrow$ & \textbf{EP} $\uparrow$ & \textbf{TTC} $\uparrow$ & \textbf{LK} $\uparrow$ & \textbf{HC} $\uparrow$ & \textbf{EC} $\uparrow$ & \textbf{EPDMS} $\uparrow$ \\
\midrule
Human Agent & --- & 100 & 100 & 99.8 & 100 & 87.4 & 100 & 100 & 98.1 & 90.1 & 90.3 \\
\midrule

Ego Status MLP & --- & 93.1 & 77.9 & 92.7 & 99.6 & 86.0 & 91.5 & 89.4 & \textbf{98.3} & 85.4 & 64.0 \\
Transfuser~\cite{transfuser} & Resnet34 & 96.9 & 89.9 & 97.8 & \underline{99.7} & 87.1 & 95.4 & 92.7 & \textbf{98.3} & \underline{87.2} & 76.7 \\
HydraMDP++ \cite{lihydra} & Resnet34 & 97.2 & \textbf{97.5} & \underline{99.4} & 99.6 & 83.1 & \underline{96.5} & 94.4 & 98.2 & 70.9 & 81.4 \\
% Artemis \cite{feng2025artemis} & Resnet34 &\textbf{98.3}& 95.1& &98.6 &99.8&81.5&97.4&96.5&100\\
DriveSuprim \cite{yao2025drivesuprim} & Resnet34 & \underline{97.5} & \underline{96.5} & \underline{99.4} & 99.6 & \textbf{88.4} & \underline{96.6} & \underline{95.5} & \textbf{98.3} & 77.0 & \underline{83.1} \\
\hline
\rowcolor{gray!20} PRIX (ours) & Resnet34 & \textbf{98.0} & 95.6 & \textbf{99.5} & \textbf{99.8} & \underline{87.4} & \textbf{97.2} &\textbf{97.1 }& \textbf{98.3}  &\textbf{ 87.6} & \textbf{84.2} \\
% \bottomrule
\end{tabular}
}
\vspace{-.2cm}
\end{table*}

% Diffusiondrive[27] ResNet34 Img. 98.0 96.0 99.5 99.8 87.7 97.1 97.2 98.3 87.6 84.3

%%%%%%%%%%%%%%%%%%%%%%%%%%%%%%%%%%%%%

%
\section{Experiments}
\label{sec:experiments}

In this section, we benchmark our method against other SOTA approaches on various datasets. When reporting baseline results, we use the scores reported by the authors, unless otherwise indicated. 

\subsection{Data and metrics}

% \paragraph{Data and metrics:}
NavSim-v1~\cite{navsim1} is a benchmark for evaluating autonomous driving agents using a non-reactive simulation where an agent plans a trajectory from initial sensor data. This approach avoids re-rendering while still enabling detailed, simulation-based analysis of the maneuver's safety and quality. Evaluation is based on the PDMS, which aggregates several metrics. It penalizes safety failures while rewarding driving performance, calculated as:

\begin{equation}
    \resizebox{\hsize}{!}{%
    $
    \text{PDMS} = \underbrace{\prod_{m \in \{\text{NC,DAC}\}} \text{score}_m}_{\text{penalties}} \times \underbrace{\frac{\sum_{w \in \{\text{EP,TTC,C}\}} \text{weight}_w \times \text{score}_w}{\sum_{w \in \{\text{EP,TTC,C}\}} \text{weight}_w}}_{\text{weighted average}}
    ,$}
    \label{eq:pdms}
\end{equation}
where penalties come from collisions (NC) and staying in the drivable area (DAC) with a weighted average of scores for progress (EP), time-to-collision (TTC), and comfort (C).

NavSim-v2~\cite{navsim2} introduces \textit{pseudo-simulation}. A planned trajectory is executed in a simulation with reactive traffic, and performance is measured by an Extended PDM Score (EPDMS). Note, NavSim-v2 is a very recent dataset and only a few approaches have been tested or adopted to it (most of them still under review).  
\begin{equation}
    \resizebox{\hsize}{!}{%
    $
    \text{EPDMS} = \underbrace{ \prod_{m \in M_{\text{pen}}} \text{filter}_{m}(\text{agent}, \text{human}) }_{ \text{penalty terms} } \cdot \underbrace{ \frac{\sum_{m \in M_{\text{avg}}} w_{m} \cdot \text{filter}_{m}(\text{agent}, \text{human})}{\sum_{m \in M_{\text{avg}}} w_{m}} }_{ \text{weighted average terms} }
    $}
    \label{eq:epdms}
\end{equation}

For nuScenes~\cite{caesar2020nuscenes}, we follow SparseDrive~\cite{sparsedrive}: stage~1 for 100 epochs followed by fine-tune stage~2 for 10 epochs using stage~1 weights. The nuScenes trajectory prediction~\cite{caesar2020nuscenes}  benchmark challenge is a popular and rich resource, where we compare our performance with a larger range of camera-only methods. Following previous works \cite{diffusiondrive,vad}, we reported average L2 error and collision rate at 3.0s horizon for a fair comparison. We compute the L2 error between the planned trajectory and the human driving trajectory, and evaluate how often we would collide with other agents on the road.

\textcolor{black}{We train on a cluster with eight NVIDIA A100 (40GB) GPUs using a per-GPU batch size of 64. For inference benchmarking (Table IV), we utilize a single \textbf{NVIDIA RTX 3090 (24GB)} GPU. To strictly simulate real-time deployment, all reported speeds use a batch size of 1 and mixed precision (FP16), encompassing the full forward pass from input tensors to final trajectory generation. We further verified these latency figures on an NVIDIA A100, observing consistent performance (variance within $\pm 0.5\%$ of the speed) on inference. On Navsim~\cite{navsim1}, we train for 100 epochs. Optimization uses AdamW with an initial learning rate of $10^{-5}$, weight decay $10^{-3}$, $(\beta_1,\beta_2)=(0.9,0.999)$, and $\epsilon=10^{-8}$. The learning rate follows a MultiStepLR schedule with milestones at epoch~70 and a decay factor $\gamma=0.1$. We apply a parameter-wise LR multiplier of 0.5 to the image encoder relative to the rest of the model. All training configurations, additional experiments, and qualitative results can be found on our \href{https://prix.netlify.app/}{project website}.}

%%%%%%%%%%%%%%%%%%%%%%%%%%%%%%%%%%555

\begin{table*}[h]
\centering
% 2. Wrap the table environment\
\resizebox{.85\textwidth}{!}{%
\begin{threeparttable} 
\caption{\small Performance comparison of different driving models for \textbf{nuScenes}. The down arrow \textbf{($\downarrow$)} indicates that \textbf{lower values are better}. Best results are in \textbf{bold}, and second best are \underline{underlined}.}
\label{tab:nuscenes}
\begin{tabular}{lcc|cccc|cccc|c}
% \toprule
\multirow{2}{*}{\textbf{Method}} & \multirow{2}{*}{\textbf{Input}} & \multirow{2}{*}{\textbf{Backbone}} & \multicolumn{4}{c|}{\textbf{L2 (m)} $\downarrow$} & \multicolumn{4}{c|}{\textbf{Collision Rate }(\%) $\downarrow$} & \multirow{2}{*}{\textbf{FPS} $\uparrow$} \\
% \cmidrule(lr){4-7} \cmidrule(lr){8-11}
& & & 1s & 2s & 3s & Avg. & 1s & 2s & 3s & Avg. & \\
\midrule
ST-P3~\cite{hu2022stp3}& Camera & EffNet-b4  & 1.33 & 2.11 & 2.90 & {2.11} & 0.23 & 0.62 & 1.27 & {0.71} & {1.6} \\
UniAD~\cite{uniad} & Camera & ResNet-101  & 0.45 & 0.70 & 1.04 & {0.73} & 0.62 & 0.58 & 0.63 & {0.61} & {1.8} \\
OccNet~\cite{liu2024fully} & Camera & ResNet-50  & 1.29 & 2.13 & 2.99 & {2.14} & 0.21 & 0.59 & 1.37 & {0.72} & {2.6} \\
VAD~\cite{vad} & Camera & ResNet-50  & 0.41 & 0.70 & 1.05 & {0.72} & 0.07 & 0.17 & 0.41 & {0.22} & {4.5} \\
SparseDrive~\cite{sparsedrive} & Camera & ResNet-50  & \underline{0.29} & \underline{0.58} & \underline{0.96} & \underline{0.61} & \underline{0.01} & \underline{0.05} & \textbf{0.18} & \underline{0.08}  & \underline{9.0} \\
% \midrule
DiffusionDrive*\tnote{\textbf{1}~~\cite{diffusiondrive}} & Camera & ResNet-50  & 0.31 & 0.62 & 1.03 & 0.65 & 0.03 & 0.06 & \underline{0.19} & 0.09 & 8.2 \\
\hline
\rowcolor{gray!20} PRIX (ours) & Camera & ResNet-50  & \textbf{0.26} & \textbf{0.53} & \textbf{0.93}& \textbf{0.57}& \textbf{0.00} & \textbf{0.04} & \textbf{0.18}& \textbf{0.07} & \textbf{11.2}  \\

\end{tabular}%

% 4. Add the tablenotes environment here
\begin{tablenotes}
    \item[*1] {\footnotesize We and other researchers were not able to reproduce results reported on nuScenes. We included the results we obtained. \url{https://github.com/hustvl/DiffusionDrive/issues/57} as well as \texttt{issues/45}. We still outperform the reported results~\cite{diffusiondrive}.}
\end{tablenotes}
\end{threeparttable}
}
\vspace{-.3cm}
\end{table*}

%%%%%%%%%%%%%%%%%%5
\subsection{Benchmarks}
\label{sec:benchmark}

By leading in the most scores and key safety metrics on Navsim-v1, Navsim-v2 and nuScenes (\cref{tab:navsim2_evaluation,tab:navsim1_results}) against the majority of the models, PRIX is an effective, well-balanced camera-only solution for autonomous navigation; qualitative examples are in \cref{fig:firstfig} (and more on the project website).

On Navsim-v1, PRIX ranks first among camera-only methods with a PDMS of 87.8; only the multimodal DiffusionDrive~\cite{diffusiondrive} scores higher overall. Notably, PRIX surpasses other models that use Camera+LiDAR inputs, such as GoalFlow~\cite{goalflow}, and attains top ranks on critical safety and performance metrics, even with adverse weather, see \cref{fig:qual_combined} (check our project website for more qualitative results). Importantly, PRIX achieves this while operating at \underline{57 FPS}, far above most baselines. Unlike HydraMDP++~\cite{lihydra} (2 frames), UniAD~\cite{uniad} ($\ge2$ frames), or VAD~\cite{chen2024vadv2} (4 frames), PRIX uses only a current frame while remaining competitive or superior across safety measures. As with all approaches, ego state information (velocity, acceleration) is provided; however, PRIX extracts more from less input, demonstrating that a camera-only, 1-frame design can still outperform methods with richer temporal or multimodal inputs. On the newer Navsim-v2 benchmark, PRIX again leads with an EPDM of 84.2 and shows robust EC performance, outperforming HydraMDP++.

PRIX also achieves SOTA performance on the nuScenes trajectory prediction challenge, outperforming all existing camera-based baselines, shown in \cref{tab:nuscenes}. In terms of average L2 error across 1s to 3s horizons, PRIX achieves the lowest value of 0.57m, surpassing the previously best DiffusionDrive (0.65 m) and SparseDrive (0.61 m). Moreover, PRIX yields the lowest collision rate at 0.07\%, with a 0.00\% collision rate at 1 second, indicating strong short-term safety. Notably, PRIX also operates at the highest inference speed (11.2 FPS), demonstrating that our model offers a superior balance of accuracy, safety, and efficiency.

\subsection{Ablations}
\label{sec:ablations}
\noindent
We further ablate different components of our model after initial design analysis in \cref{sec:design}. All ablations are done on Navsim-v1. More ablations can be found on our \href{https://prix.netlify.app/}{project website}.
\noindent
\textbf{Loss influence:} We demonstrate the progressive benefit of each auxiliary loss. We show that using only planning as supervision is not enough. The baseline model, using only the planning loss ($\mathcal{L}_{\text{plan}}$), scores 70.4 on PDMS. Adding tasks responsible for environment understanding, such as agent detection and classification, plus semantic segmentation, successively boosts the score as shown in \cref{tab:loss_ablation}. That confirms that the planner's performance is directly coupled with the quality of the features, which learn a semantically rich representation of the scene through these auxiliary tasks.

\begin{table}[h!] \centering \caption{Contribution of each loss component.} \label{tab:loss_ablation} 
\renewcommand{\arraystretch}{0.8} % Default value: 1
\setlength{\tabcolsep}{5pt}
% \vspace{-.3cm}
% \resizebox{!}{.9\textwidth}{
\begin{tabular}{c | cccc | c}
% \toprule 
\textbf{Exp. \#} & $\mathcal{L}_{\text{plan}}$ & $\mathcal{L}_{\text{box}}$ & $\mathcal{L}_{\text{sem}}$ & $\mathcal{L}_{\text{cls}}$ & \textbf{PDMS $\uparrow$}
\\ \toprule
1 & \checkmark &&&& 70.4
\\ 2 & \checkmark &\checkmark &&& 82.3
\\ 2 & \checkmark & &\checkmark&& 85.7
\\ 3 & \checkmark &\checkmark& \checkmark && 86.9
% \\ 4 & \checkmark &\checkmark & \checkmark & \checkmark & 86.4 
\\ \midrule 4 (Full) & \checkmark & \checkmark & \checkmark & \checkmark & \textbf{87.8} 
% \\ \bottomrule 
\end{tabular} 
\vspace{-.2cm}
\end{table} 

\noindent
\textbf{Different Planners}: 
Results in~\cref{tab:planners} affirm our core hypothesis that visual feature extractor is the most critical component. While our top-performing diffusion planner is also the slowest at 57.0 FPS, a simple MLP head is highly competitive. This strong performance from a minimal planner proves the richness of the learned visual representation. A clear trade-off exists: for applications requiring higher speed, the diffusion head can be swapped for much faster alternatives, like the MLP or the second-best LSTM, with only a minor compromise in accuracy. This confirms that foundational heavy lifting is handled by the visual encoder.
\begin{table}[h!]
\centering
\caption{Planners comparison, all models use ResNet34.}
\label{tab:planners}
\vspace{-.2cm}
\resizebox{\columnwidth}{!}{%
\begin{tabular}{llccc}
% \toprule
\textbf{Model} & \textbf{Planner} & \textbf{PDMS} $\uparrow$ &  \textbf{Params} $\downarrow$ & \textbf{FPS} $\uparrow$ \\
\midrule
\rowcolor{gray!20} PRIX (baseline) & Diffusion & \textbf{87.8} & 37M & 57.0\\
PRIX-mlp & MLP & 85.1 &  \textbf{33M }& \textbf{65.3}  \\
PRIX-t & Transformer & 85.4 &35M& 62.8 \\
PRIX-ls & LSTM & 86.7 & 34M & 63.4\\

\end{tabular}%
}
\end{table}

\noindent
\textbf{Limitation and future work}
While PRIX achieves great performance on every popular end-to-end driving benchmark, its camera-only nature makes it vulnerable to adverse weather, occlusions, and sensor failure or decalibration. Future work can enhance robustness through two main avenues. First, self-supervised pre-training on large, unlabeled datasets could help the backbone learn more resilient features~\cite{seal,govindarajan2025cleverdistiller}. Second, incorporating control~\cite{nyberg2021risk} or rule-based approaches~\cite{zheng2025diffusion} could better manage uncertainties and improve safety in challenging scenarios. While PRIX performs well on adverse weather or partially missing data, it does not generalize between different sensor setups and would have to be retrain for each. Generalization is something we will explore in future work. Additionally, recent works argue that good representation and planning can be learnt and with supervision based on future frame prediction \cite{li2025navigation}, which is worth trying in the future.

\section{Conclusions}
\noindent
We introduce PRIX, an efficient camera-only driving model that outperforms other published vision-based methods and rivals SOTA multimodal systems. While LiDAR remains important for robustness, we show that high performance is achievable with vision alone, demonstrating that leveraging rich camera features for planning is a viable alternative to BEV and multimodal approaches, and establishing a new benchmark for efficient, vision-based autonomous driving.

% \linebreak

% \noindent \textbf{Acknowledgment} This work was partially supported by the Wallenberg AI, Autonomous Systems and Software Program (WASP) funded by the Knut and Alice Wallenberg Foundation.

\nocite{Dan_Flaticon_F1}
{\footnotesize
\bibliographystyle{IEEEtran} %{ieee_fullname}
\bibliography{references}

@STRING{CVPR="CVPR"}

@STRING{ECCV="ECCV"}

@STRING{ICCV="ICCV"}

@STRING{TPAMI="TPAMI"}

@STRING{IROS="IROS"}

@STRING{ICRA="ICRA"}

@STRING{RAL="RAL"}

@STRING{ICLR="ICLR"}

@STRING{BMVC="BMVC"}

@STRING{COMPUTER="Computer"}

@STRING{IV="IV"}

@STRING{FUSION="FUSION"}

@STRING{NIPS="NeurIPS"}

@STRING{IROS="iROS"}

@STRING{RAL="RA-L"}

@STRING{IV="IEEE Intelligent Vehicles Symposium (IV)"}

@STRING{FUSION="Proceedings of the IEEE International Conference on Information Fusion (FUSION)"}

@inproceedings{uniad,
  title={Planning-oriented autonomous driving},
  author={Hu, Yihan and Yang, Jiazhi and Chen, Li and Li, Keyu and Sima, Chonghao and Zhu, Xizhou and Chai, Siqi and Du, Senyao and Lin, Tianwei and Wang, Wenhai and others},
  booktitle=CVPR,
  pages={17853--17862},
  year={2023}
}

@inproceedings{goalflow,
  title={Goalflow: Goal-driven flow matching for multimodal trajectories generation in end-to-end autonomous driving},
  author={Xing, Zebin and Zhang, Xingyu and Hu, Yang and Jiang, Bo and He, Tong and Zhang, Qian and Long, Xiaoxiao and Yin, Wei},
  booktitle=CVPR,
  pages={1602--1611},
  year={2025}
}

@article{lihydra,
  title={Hydra-MDP++: Advancing End-to-End Driving via Hydra-Distillation with Expert-Guided Decision Analysis},
  author={Li, Kailin and Li, Zhenxin and Lan, Shiyi and Liu, Jiayi and Xie, Yuan and Wu, Zuxuan and Yu, Zhiding and Alvarez, Jose M and others},
  journal=CVPR,
  year={2025}
}

@inproceedings{hu2022stp3,
 title={ST-P3: End-to-end Vision-based Autonomous Driving via Spatial-Temporal Feature Learning}, 
 author={Shengchao Hu and Li Chen and Penghao Wu and Hongyang Li and Junchi Yan and Dacheng Tao},
 booktitle=ECCV,
 year={2022}
}

@article{wozniak2023toward,
  title={Toward a robust sensor fusion step for 3D object detection on corrupted data},
  author={Wozniak, Maciej K and K{\aa}refj{\"a}rd, Viktor and Thiel, Marko and Jensfelt, Patric},
  journal=RAL,
  volume={8},
  number={11},
  pages={7018--7025},
  year={2023},
  publisher={IEEE}
}

@article{li2025navigation,
  title={Navigation-guided sparse scene representation for end-to-end autonomous driving},
  author={Li, Peidong and Cui, Dixiao},
  journal=ICLR,
  year={2025}
}

@inproceedings{Dan_Flaticon_F1,
  author       = {Darius Dan},
  title        = {Formula 1 Icons},
  organization = {Flaticon},
  booktitle          = {https://www.flaticon.com/free-icons/formula-1},
  urldate      = {2025-07-17}
}

@inproceedings{nyberg2021risk,
  title={Risk-aware motion planning for autonomous vehicles with safety specifications},
  author={Nyberg, Truls and Pek, Christian and Dal Col, Laura and Nor{\'e}n, Christoffer and Tumova, Jana},
  booktitle={IV},
  pages={1016--1023},
  year={2021},
  organization={IEEE}
}

@article{govindarajan2025cleverdistiller,
  title={CleverDistiller: Simple and Spatially Consistent Cross-modal Distillation},
  author={Govindarajan, Hariprasath and Wozniak, Maciej K and Klingner, Marvin and Maurice, Camille and Kiran, B Ravi and Yogamani, Senthil},
  journal=BMVC,
  year={2025}
}

@article{chen2024end,
  title={End-to-end autonomous driving: Challenges and frontiers},
  author={Chen, Li and Wu, Penghao and Chitta, Kashyap and Jaeger, Bernhard and Geiger, Andreas and Li, Hongyang},
  journal=TPAMI,
  year={2024},
  publisher={IEEE}
}

@misc{liu2024fully,
    title={Fully Sparse 3D Occupancy Prediction}, 
    author={Haisong Liu and Yang Chen and Haiguang Wang and Zetong Yang and Tianyu Li and Jia Zeng and Li Chen and Hongyang Li and Limin Wang},
    booktitle=ECCV,
    pages={18428--18435},
    year={2024},
}

@inproceedings{xu2024towards,
  title={Towards motion forecasting with real-world perception inputs: Are end-to-end approaches competitive?},
  author={Xu, Yihong and Chambon, Lo{\"\i}ck and Zablocki, {\'E}loi and Chen, Micka{\"e}l and Alahi, Alexandre and Cord, Matthieu and P{\'e}rez, Patrick},
  booktitle=ICRA,
  pages={18428--18435},
  year={2024},
  organization={IEEE}
}

@article{gao2025rad,
  title={Rad: Training an end-to-end driving policy via large-scale 3dgs-based reinforcement learning},
  author={Gao, Hao and Chen, Shaoyu and Jiang, Bo and Liao, Bencheng and Shi, Yiang and Guo, Xiaoyang and Pu, Yuechuan and Yin, Haoran and Li, Xiangyu and Zhang, Xinbang and others},
  journal={arXiv preprint arXiv:2502.13144},
  year={2025}
}

@article{feng2025artemis,
  title={Artemis: Autoregressive end-to-end trajectory planning with mixture of experts for autonomous driving},
  author={Feng, Renju and Xi, Ning and Chu, Duanfeng and Wang, Rukang and Deng, Zejian and Wang, Anzheng and Lu, Liping and Wang, Jinxiang and Huang, Yanjun},
  journal={arXiv preprint arXiv:2504.19580},
  year={2025}
}

@inproceedings{hess2025splatad,
  title={Splatad: Real-time lidar and camera rendering with 3d gaussian splatting for autonomous driving},
  author={Hess, Georg and Lindstr{\"o}m, Carl and Fatemi, Maryam and Petersson, Christoffer and Svensson, Lennart},
  booktitle=CVPR,
  pages={11982--11992},
  year={2025}
}

@inproceedings{worldnavigation,
  title={Navigation world models},
  author={Bar, Amir and Zhou, Gaoyue and Tran, Danny and Darrell, Trevor and LeCun, Yann},
  booktitle=CVPR,
  pages={15791--15801},
  year={2025}
}

@inproceedings{lss,
  title={Lift, splat, shoot: Encoding images from arbitrary camera rigs by implicitly unprojecting to 3d},
  author={Philion, Jonah and Fidler, Sanja},
  booktitle=ECCV,
  pages={194--210},
  year={2020},
  organization={Springer}
}

@inproceedings{navsim1,
	title = {NAVSIM: Data-Driven Non-Reactive Autonomous Vehicle Simulation and Benchmarking},
	author = {Daniel Dauner and Marcel Hallgarten and Tianyu Li and Xinshuo Weng and Zhiyu Huang and Zetong Yang and Hongyang Li and Igor Gilitschenski and Boris Ivanovic and Marco Pavone and Andreas Geiger and Kashyap Chitta},
	booktitle = NIPS,
	year = {2024},
}

@article{navsim2,
	title={Pseudo-Simulation for Autonomous Driving}, 
        author={Wei Cao and Marcel Hallgarten and Tianyu Li and Daniel Dauner and Xunjiang Gu and Caojun Wang and Yakov Miron and Marco Aiello and Hongyang Li and Igor Gilitschenski and Boris Ivanovic and Marco Pavone and Andreas Geiger and Kashyap Chitta},
	journal = {CoRL},
	year = {2025},
}

@article{transdiffuser,
  title={TransDiffuser: End-to-end Trajectory Generation with Decorrelated Multi-modal Representation for Autonomous Driving},
  author={Jiang, Xuefeng and Ma, Yuan and Li, Pengxiang and Xu, Leimeng and Wen, Xin and Zhan, Kun and Xia, Zhongpu and Jia, Peng and Lang, XianPeng and Sun, Sheng},
  journal={arXiv preprint arXiv:2505.09315},
  year={2025}
}

@article{yuan2024drama,
  title={Drama: An efficient end-to-end motion planner for autonomous driving with mamba},
  author={Yuan, Chengran and Zhang, Zhanqi and Sun, Jiawei and Sun, Shuo and Huang, Zefan and Lee, Christina Dao Wen and Li, Dongen and Han, Yuhang and Wong, Anthony and Tee, Keng Peng and others},
  journal={arXiv preprint arXiv:2408.03601},
  year={2024}
}

@article{zheng2025diffusion,
  title={Diffusion-based planning for autonomous driving with flexible guidance},
  author={Zheng, Yinan and Liang, Ruiming and Zheng, Kexin and Zheng, Jinliang and Mao, Liyuan and Li, Jianxiong and Gu, Weihao and Ai, Rui and Li, Shengbo Eben and Zhan, Xianyuan and others},
  journal=ICLR,
  year={2025}
}

@article{li2024hydra,
  title={Hydra-mdp: End-to-end multimodal planning with multi-target hydra-distillation},
  author={Li, Zhenxin and Li, Kailin and Wang, Shihao and Lan, Shiyi and Yu, Zhiding and Ji, Yishen and Li, Zhiqi and Zhu, Ziyue and Kautz, Jan and Wu, Zuxuan and others},
  journal={arXiv preprint arXiv:2406.06978},
  year={2024}
}

@article{qiao2025lightemma,
  title={Lightemma: Lightweight end-to-end multimodal model for autonomous driving},
  author={Qiao, Zhijie and Li, Haowei and Cao, Zhong and Liu, Henry X},
  journal={arXiv preprint arXiv:2505.00284},
  year={2025}
}

@article{hegde2025distilling,
  title={Distilling Multi-modal Large Language Models for Autonomous Driving},
  author={Hegde, Deepti and Yasarla, Rajeev and Cai, Hong and Han, Shizhong and Bhattacharyya, Apratim and Mahajan, Shweta and Liu, Litian and Garrepalli, Risheek and Patel, Vishal M and Porikli, Fatih},
  journal=CVPR,
  year={2025}
}

@article{jiang2025diffvla,
  title={Diffvla: Vision-language guided diffusion planning for autonomous driving},
  author={Jiang, Anqing and Gao, Yu and Sun, Zhigang and Wang, Yiru and Wang, Jijun and Chai, Jinghao and Cao, Qian and Heng, Yuweng and Jiang, Hao and Zhang, Zongzheng and others},
  journal={arXiv preprint arXiv:2505.19381},
  year={2025}
}

@inproceedings{chen2025solve,
  title={SOLVE: Synergy of Language-Vision and End-to-End Networks for Autonomous Driving},
  author={Chen, Xuesong and Huang, Linjiang and Ma, Tao and Fang, Rongyao and Shi, Shaoshuai and Li, Hongsheng},
  booktitle=CVPR,
  pages={12068--12077},
  year={2025}
}

@inproceedings{paul2024lego,
  title={LeGo-Drive: Language-enhanced Goal-oriented Closed-Loop End-to-End Autonomous Driving},
  author={Paul, Pranjal and Garg, Anant and Choudhary, Tushar and Singh, Arun Kumar and Krishna, K Madhava},
  booktitle=IROS,
  pages={10020--10026},
  year={2024},
  organization={IEEE}
}

@article{ivanovic2025efficient,
  title={Efficient Multi-Camera Tokenization with Triplanes for End-to-End Driving},
  author={Ivanovic, Boris and Saltori, Cristiano and You, Yurong and Wang, Yan and Luo, Wenjie and Pavone, Marco},
  journal={arXiv preprint arXiv:2506.12251},
  year={2025}
}

@article{li2025generalized,
  title={Generalized Trajectory Scoring for End-to-end Multimodal Planning},
  author={Li, Zhenxin and Yao, Wenhao and Wang, Zi and Sun, Xinglong and Chen, Joshua and Chang, Nadine and Shen, Maying and Wu, Zuxuan and Lan, Shiyi and Alvarez, Jose M},
  journal={arXiv preprint arXiv:2506.06664},
  year={2025}
}

@article{yao2025drivesuprim,
  title={DriveSuprim: Towards Precise Trajectory Selection for End-to-End Planning},
  author={Yao, Wenhao and Li, Zhenxin and Lan, Shiyi and Wang, Zi and Sun, Xinglong and Alvarez, Jose M and Wu, Zuxuan},
  journal={arXiv preprint arXiv:2506.06659},
  year={2025}
}

@article{yasarla2025roca,
  title={RoCA: Robust Cross-Domain End-to-End Autonomous Driving},
  author={Yasarla, Rajeev and Han, Shizhong and Cheng, Hsin-Pai and Liu, Litian and Mahajan, Shweta and Bhattacharyya, Apratim and Shi, Yunxiao and Garrepalli, Risheek and Cai, Hong and Porikli, Fatih},
  journal={arXiv preprint arXiv:2506.10145},
  year={2025}
}

@inproceedings{chen2024ppad,
  title={Ppad: Iterative interactions of prediction and planning for end-to-end autonomous driving},
  author={Chen, Zhili and Ye, Maosheng and Xu, Shuangjie and Cao, Tongyi and Chen, Qifeng},
  booktitle=ECCV,
  pages={239--256},
  year={2024},
  organization={Springer}
}

@inproceedings{diffusiondrive,
  title={Diffusiondrive: Truncated diffusion model for end-to-end autonomous driving},
  author={Liao, Bencheng and Chen, Shaoyu and Yin, Haoran and Jiang, Bo and Wang, Cheng and Yan, Sixu and Zhang, Xinbang and Li, Xiangyu and Zhang, Ying and Zhang, Qian and others},
  booktitle=CVPR,
  pages={12037--12047},
  year={2025}
}

@article{sparsedrive,
  title={Sparsedrive: End-to-end autonomous driving via sparse scene representation},
  author={Sun, Wenchao and Lin, Xuewu and Shi, Yining and Zhang, Chuang and Wu, Haoran and Zheng, Sifa},
  journal=ICRA,
  year={2025}
}

@inproceedings{vad,
  title={Vad: Vectorized scene representation for efficient autonomous driving},
  author={Jiang, Bo and Chen, Shaoyu and Xu, Qing and Liao, Bencheng and Chen, Jiajie and Zhou, Helong and Zhang, Qian and Liu, Wenyu and Huang, Chang and Wang, Xinggang},
  booktitle=ICCV,
  pages={8340--8350},
  year={2023}
}

@article{transfuser,
  title={Transfuser: Imitation with transformer-based sensor fusion for autonomous driving},
  author={Chitta, Kashyap and Prakash, Aditya and Jaeger, Bernhard and Yu, Zehao and Renz, Katrin and Geiger, Andreas},
  journal=TPAMI,
  volume={45},
  number={11},
  pages={12878--12895},
  year={2022},
  publisher={IEEE}
}

@inproceedings{hu2018squeeze,
  title={Squeeze-and-excitation networks},
  author={Hu, Jie and Shen, Li and Sun, Gang},
  booktitle={Proceedings of the IEEE conference on computer vision and pattern recognition},
  pages={7132--7141},
  year={2018}
}

@inproceedings{woo2018cbam,
  title={Cbam: Convolutional block attention module},
  author={Woo, Sanghyun and Park, Jongchan and Lee, Joon-Young and Kweon, In So},
  booktitle={Proceedings of the European conference on computer vision (ECCV)},
  pages={3--19},
  year={2018}
}

@article{chen2024vadv2,
  title={Vadv2: End-to-end vectorized autonomous driving via probabilistic planning},
  author={Chen, Shaoyu and Jiang, Bo and Gao, Hao and Liao, Bencheng and Xu, Qing and Zhang, Qian and Huang, Chang and Liu, Wenyu and Wang, Xinggang},
  journal={arXiv preprint arXiv:2402.13243},
  year={2024}
}

@inproceedings{paradrive,
  title={Para-drive: Parallelized architecture for real-time autonomous driving},
  author={Weng, Xinshuo and Ivanovic, Boris and Wang, Yan and Wang, Yue and Pavone, Marco},
  booktitle=CVPR,
  pages={15449--15458},
  year={2024}
}

@inproceedings{tonderski2024neurad,
  title={Neurad: Neural rendering for autonomous driving},
  author={Tonderski, Adam and Lindstr{\"o}m, Carl and Hess, Georg and Ljungbergh, William and Svensson, Lennart and Petersson, Christoffer},
  booktitle=CVPR,
  pages={14895--14904},
  year={2024}
}

@inproceedings{caesar2020nuscenes,
  title={nuscenes: A multimodal dataset for autonomous driving},
  author={Caesar, Holger and Bankiti, Varun and Lang, Alex H and Vora, Sourabh and Liong, Venice Erin and Xu, Qiang and Krishnan, Anush and Pan, Yu and Baldan, Giancarlo and Beijbom, Oscar},
  booktitle=CVPR,
  pages={11621--11631},
  year={2020}
}

@article{seal,
  title={Segment any point cloud sequences by distilling vision foundation models},
  author={Liu, Youquan and Kong, Lingdong and Cen, Jun and Chen, Runnan and Zhang, Wenwei and Pan, Liang and Chen, Kai and Liu, Ziwei},
  journal=NIPS,
  volume={36},
  year={2024}
}
}
\end{document}